%% file: ETPP_SAC_2020.tex
\documentclass[sigconf]{acmart}

\setcopyright{acmcopyright}

\usepackage{subfigure}
\usepackage{enumerate}
\usepackage{amssymb}
\usepackage{amsthm}
\usepackage{graphicx} 
\usepackage{subfigure}

\usepackage{mathrsfs}
\usepackage{amsmath} 
\usepackage{hyperref} 
\usepackage[ruled]{algorithm2e}
\usepackage{color}
\usepackage{array}
\usepackage{mathtools}
\usepackage{bbm, dsfont}

\acmDOI{xx.xxx/xxx_x}

\acmISBN{978-1-4503-6866-7/20/03}

\acmConference[SAC'20]{ACM SAC Conference}{March 30-April 3, 2020}{Brno, Czech Republic} 
\acmYear{2020}
\copyrightyear{2020}

\acmArticle{4}
\acmPrice{15.00}

\begin{document}
\title{Event Ticket Price Prediction with Deep Neural Network \\ on Spatial-Temporal Sparse Data}

\author{Fei Huang}
\authornote{Most of this work was done when the first author was working in GE Digital.}
\affiliation{%
 \institution{Microsoft}
 \streetaddress{555 110th Ave NE}
 \city{Redmond} 
 \state{WA} 
 \postcode{98052}
}
\email{fhfeihuangfh@gmail.com}

\author{Hao Huang}
\affiliation{%
 \institution{GE Global Research}
 \streetaddress{2623 Camino Ramon}
 \city{San Ramon} 
 \state{CA} 
 \postcode{94583}
}
\email{haohuanghw@gmail.com}

\begin{abstract}
Event ticket price prediction is important to marketing strategy for any sports team or musical ensemble. An accurate prediction model can help the marketing team to make promotion plan more effectively and efficiently. However, given all the historical transaction records, it is challenging to predict the sale price of the remaining seats at any future timestamp, not only because that the sale price is relevant to a lot of features (seat locations, date-to-event of the transaction, event date, team performance, etc.), but also because of the temporal and spatial sparsity in the dataset. For a game/concert, the ticket selling price of one seat is only observable once at the time of sale. Furthermore, some seats may not even be purchased (therefore no record available). In fact, data sparsity is commonly encountered in many prediction problems. Here, we propose a bi-level optimizing deep neural network to address the curse of spatio-temporal sparsity. Specifically, we introduce coarsening and refining layers, and design a bi-level loss function to integrate different level of loss for better prediction accuracy. Our model can discover the interrelations among ticket sale price, seat locations, selling time, event information, etc. Experiments show that our proposed model outperforms other benchmark methods in real-world ticket selling price prediction.  
\end{abstract}

%
%
\copyrightyear{2020} 
\acmYear{2020} 
\setcopyright{acmlicensed}
\acmConference[SAC '20]{The 35th ACM/SIGAPP Symposium on Applied Computing}{March 30-April 3, 2020}{Brno, Czech Republic}
\acmBooktitle{The 35th ACM/SIGAPP Symposium on Applied Computing (SAC '20), March 30-April 3, 2020, Brno, Czech Republic}
\acmPrice{15.00}
\acmDOI{10.1145/3341105.3373840}
\acmISBN{978-1-4503-6866-7/20/03}



\maketitle

\input{Introduction}
\input{ProposedModel}

\input{Discussion}

\input{Experiment}

\input{Conclusion}

\bibliographystyle{ACM-Reference-Format}
\bibliography{ref}

\end{document}

%% file: Introduction.tex
\begin{figure*}[!h]
\centering
\includegraphics[width=1.0\linewidth]{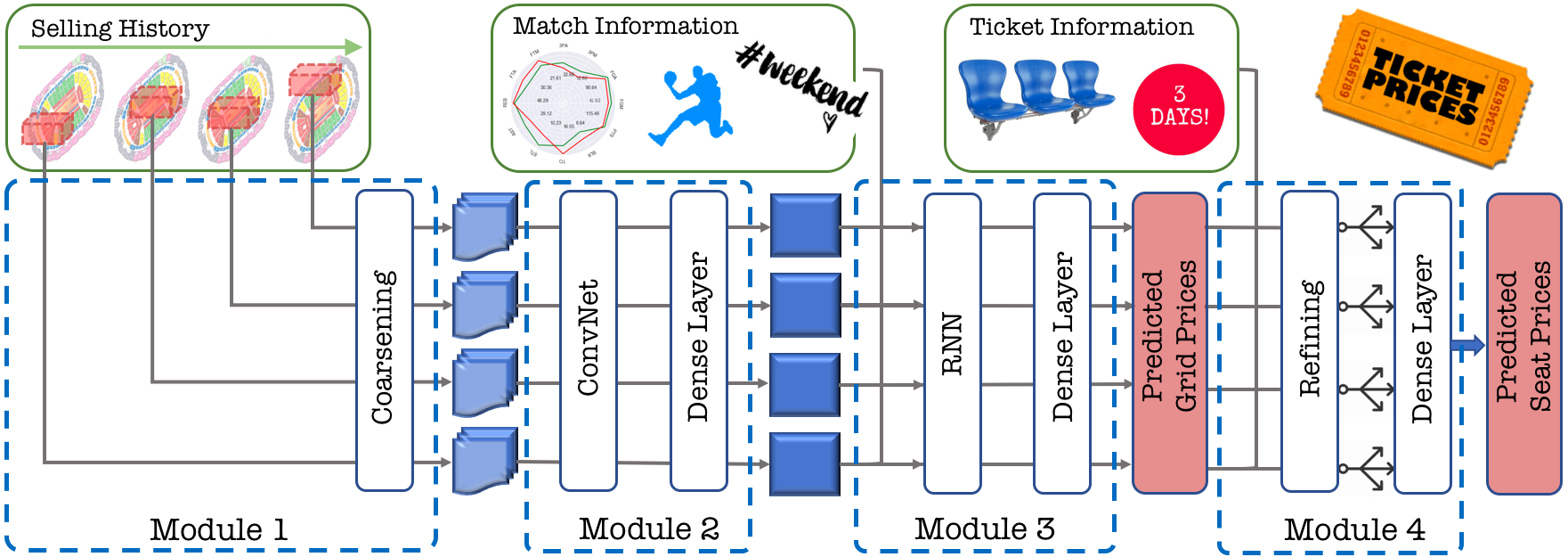}
 \caption{Our Event Ticket Price Prediction (ETPP) Model. It consists of four modules. Module $1$ is to coarsen the input to mitigate the data sparsity. Module $2$ is to learn the spatial dependency via ConvNet, while temporal dependency is explored with event (match) information by RNN in Module $3$. Finally Module $4$ projects the coarsened prediction to original seat level, and calculates the loss by a specially designed bi-level loss function. }
 \label{fig:framework}
\end{figure*}

\section{Introduction}
\label{sec:intro}

Promotional campaign, a series of advertisements using various marketing tools to promote an event (such as a concert or sports match), plays an important role in reaching out to potential customers. It can make the event better known in the relevant circles and promote event ticket sales. 
A successful advertising campaign cannot be executed in a vacuum and therefore, brands and businesses should carefully launch the campaign at the right moment. This is important because timing is crucial to the success of any public campaign. Among all the factors, predicting future ticket sale price of the targeted event is one of the most important ones, since it will lead to profit maximization if a campaign run right before the ticket selling price reaching its peak. So the team can decide when to get the most out of limited resources to promote the event in advance.  

However, predicting ticket sale price for a concert or sports match is nontrivial, and it is different from traditional ticket price prediction such as airline price prediction ~\cite{mottini2017deep,lu2017machine,groves2013optimal}. Airline ticket price prediction focus on predicting the average price of a flight, or a particular class (such as Economy Class), which doesn't distinguish the spatial difference between seats. But match/concert ticket price is relevant to seat location since price can vary a lot for different seats, and it is important to predict each seat's selling price so that the campaign plan can be more oriented to specific groups of potential customers. Therefore predicting event ticket selling price is a spatio-temporal analysis problem. 

Spatio-temporal prediction has become a popular problem for a wide range of settings ranging from weather/climate forecasting \cite{liu2018distribution,xu2018muscat}, video prediction \cite{oh2015action,singh2017online}, traffic flow forecasting \cite{zhang2017deep,yao2019revisiting,yu2017spatio,yu2017spatiotemporal} to taxi demanding prediction \cite{yao2018deep,ke2017short}. 
In spatio-temporal prediction, measurements are often taken over space and time, while the model is required to predict future target value at any location and time. 

Different from the traditional spatio-temporal prediction problem, our problem has the following challenges:
\begin{enumerate}
\item The available records of ticket transaction are sparse in both space and time, in the sense that for any historical event i) the selling price for one seat was only observable once at the time of sale, ii) it is not unusual that some seats didn't get sold, iii) the distribution of the transaction time is very uneven.    
\item Event ticket selling price is relevant to multiple types of information including temporal (transaction time), spatial (seat locations) and event information (event date, number of team stars, team's recent performance, etc.). 
\end{enumerate}
Unfortunately, there is few research that aim at predicting the event ticket selling price to solve these challenges.        

In this paper, we target to solve the above challenges by proposing \textbf{E}vent \textbf{T}icket \textbf{P}rice \textbf{P}rediction (\textit{\textbf{ETPP}}). The framework is shown in Figure \ref{fig:framework}. Our research has the following contributions:
\begin{enumerate}
\item We resolve the data sparsity problem by proposing data coarsening and refining layers \textit{on both spatial and temporal dimension}. The coarsening layer converts the original sparse input to a coarse but much less sparse format, so to avoid the zero weighting problem of deep neural network on sparse data. The refining layer maps the coarse resolution back to original input level, so that the predicted value can be directly compared with ground truth.   
\item Integrating the predicted values on both coarsened and original level, we propose a bi-level optimization method, and design a bi-level loss function that involves two levels of predicted output, which provides better prediction accuracy. 
\item Our ETPP model systematically combines data coarsening and refining layers, spatial and temporal modeling and a special loss function. It considers multiple types of input (ticket transaction history, event information, ticket information, etc.), and outputs accurate prediction results on a real world ticket transaction dataset.   
\end{enumerate}

%% file: ProposedModel.tex
\section{Our ETPP Framework}
\label{sec:framework}
In this section, we first define the problem formally, and then elaborate our ETPP in details. As shown in Figure \ref{fig:framework}, ETPP consists of four modules. Module $1$ converts the input data into coarse format to mitigate the data sparsity. Then convolutional net (ConvNet) and fully connected layers are applied in Module $2$ to extract the spatial dependency. Module $3$ focuses on modeling temporal dependency by recurrent neural network (RNN). Finally, Module $4$ projects the data resolution back to the original level and calculates the final loss function. 

\subsection{Problem Statement}
\label{sec:problem_statement}

Given the historical records of ticket transactions, we want to predict the seats' sale price at any time before the event. Each transaction record includes the selling price, the transaction time and the location of the corresponding seat, and the corresponding event information (such as event date, team's recent performance, number of team stars, etc.). 

Suppose there are $K$ historical events and the location has $n$ number of seats \footnote{We assume that all the events were held at the same location, which can be a stadium (for sports match) or hall (for concert).}. For any event $k\in 1,2,...K$,  we denote $p_k{(t, r, c)}$ as the ticket price of the seat at the $r$-th row and the $c$-th column, with date-to-event $t$ as the transaction time.  We use $P_{k,t}  \in \mathbb{R}^{n \times 1}$ to denote all the seat prices at date-to-event $t$ (nans for certain seats if there were no corresponding transactions at $t$).

Our goal is to design a model to predict $P_{K+c,t}$ ($c=1,2,...$) ahead given all the available historical transactions $P_{k,*}$ and event information $\mathcal{I}_k$ (event date, team's recent performance, number of team stars, etc.) where $k=1,2,...,K$. 

\begin{figure}[!h]
\centering
\includegraphics[width=0.9\linewidth]{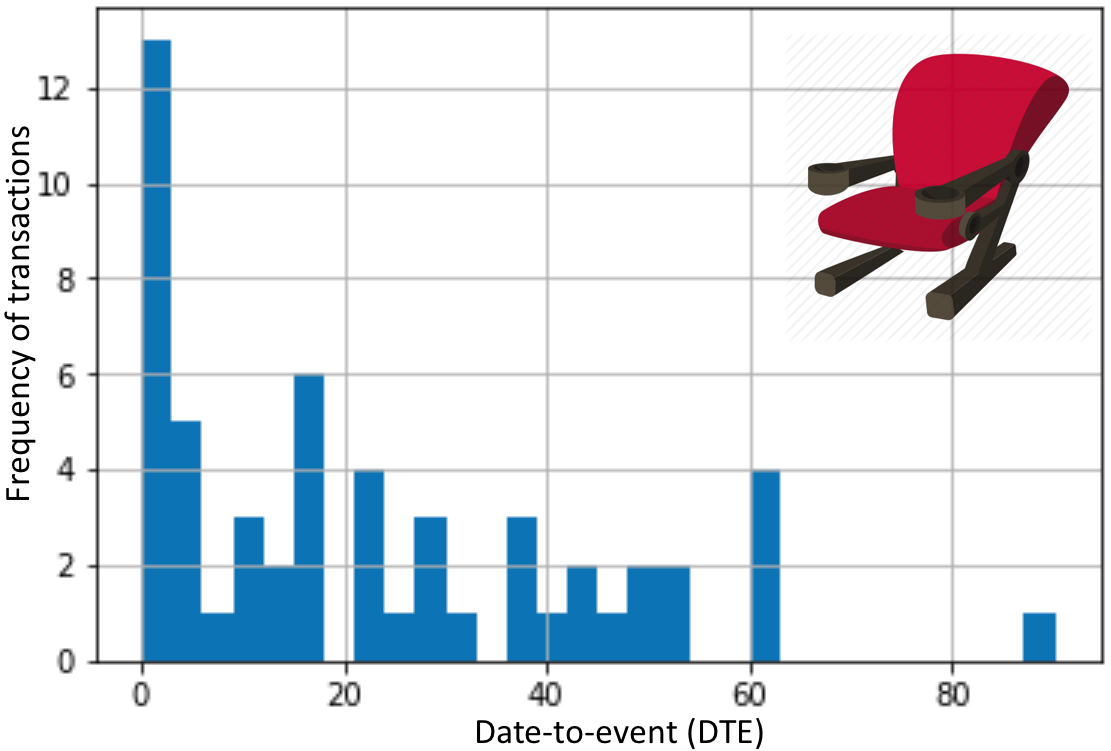}
 \caption{Histogram across date-to-event of a certain seat's transactions for $246$ historical events. Obviously, the distribution is quite uneven, where most transactions happened right before the event date, and very few happened three months ahead.}
 \label{fig:DTE_seat}
\end{figure}

However, the historical record of ticket transaction is usually very sparse in time and space, and most of $P_{k,t}$ are with empty values because few transactions happened at $t$, or some seats didn't even get sold for that event. Figure \ref{fig:DTE_seat} shows the transaction histogram across date-to-event of a certain seat for $246$ historical events. Obviously, the distribution is quite uneven, where most transactions happened right before the event date, and very few happened three months ahead. 
It is also not difficult to see that $P_{k,t}$ could be extremely sparse at certain date-to-event $t$. On the other hand, any seat of an event can be only purchased once (assume there is no cancellation or reselling). Therefore, the historical data is very sparse in both time (date-to-event) and space (seat), which makes traditional regression methods inapplicable. 
And existing deep neural network also suffer from zero weighting problem on such sparse data.
To solve this problem, our first step is to propose a coarsening layer on both temporal and spatial dimensions as described in Section \ref{sec:module1}.

\begin{figure}[!h]
\centering
\includegraphics[width=1.0\linewidth]{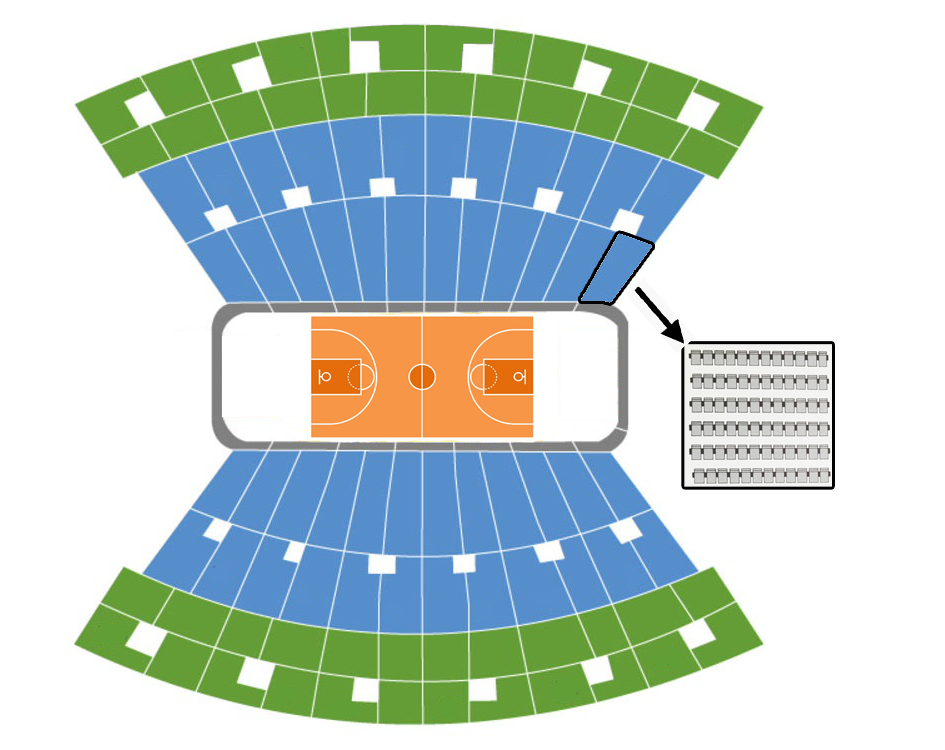}
 \caption{Module $1$ splits the auditorium into $m$ non-overlapping grids. Each grid is later expanded back to seat-level resolution in Module $4$.}
 \label{fig:auditorium-grid}
\end{figure}

\subsection{Module 1: Data Coarsening}
\label{sec:module1}
To solve the data sparsity problem, here we introduce a data coarsening layer on both spatial and temporal dimension. Assuming the event location is an auditorium, which has $n$ seats in total, we first deal with spatial sparsity by splitting $n$ seats into $m$ non-overlapping \textbf{grids} $GS = \{gs_1,\allowbreak gs_2,\allowbreak...,\allowbreak gs_m\}$.  Each grid $gs_i$ ($i = 1,2,...,m$) consists of a few seats $gs_i = \{  gs_{i(1,1)},\allowbreak gs_{i(1,2)},\allowbreak ...,\allowbreak gs_{i(2,1)},\allowbreak  gs_{i(2,2)},\allowbreak ...   \}$ where $gs_{i(i_r,i_c)}$ denotes the seat in $i_r$-th row and $i_c$-th column in grid $gs_i$. This  is illustrated in Figure \ref{fig:auditorium-grid}.

\begin{figure}[!h]
	\centering
	\subfigure[Transactions histogram on DTE of all the available records.]{\includegraphics[width=0.8\linewidth]{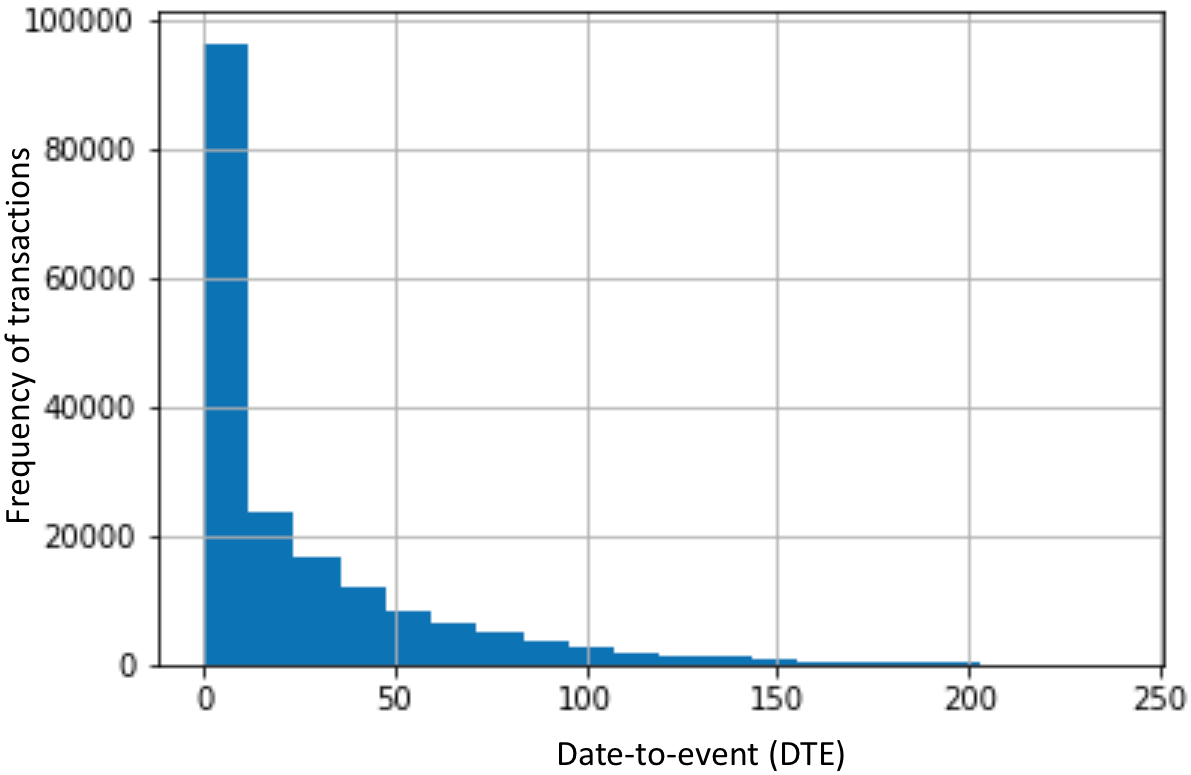} \label{fig:DTE_all}}
	\subfigure[Transactions histogram after converting to Log(DTE+1)]{\includegraphics[width=0.8\linewidth]{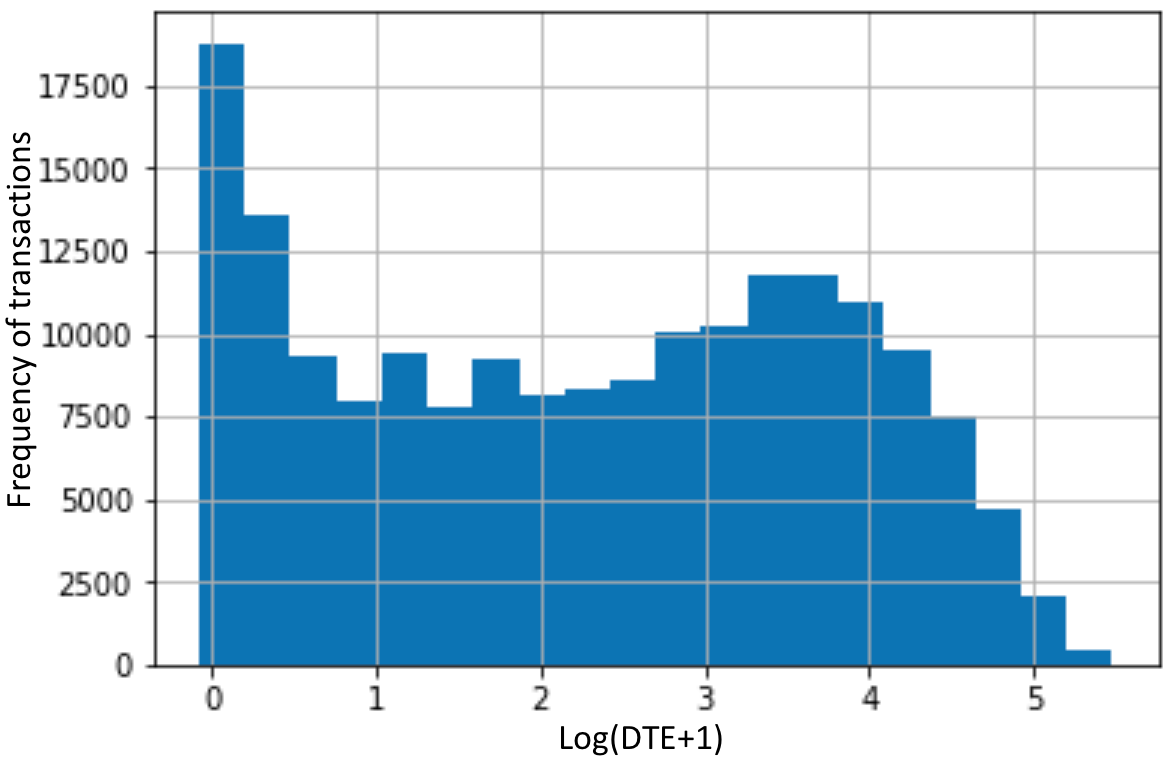} \label{fig:DTE_all_log}}
	\caption{Temporal conversion.}
	\label{fig:DTE_log_setting}
\end{figure}

In temporal dimension, the data availability is not evenly distributed. Figure \ref{fig:DTE_all} shows the date-to-event (DTE)  histogram of all the transactions in $246$ historical events. Obviously it is not reasonable to coarsen temporal data with one fixed-size. We notice from Figure \ref{fig:DTE_all} that ticket transactions are rare when DTE is large, and the frequency increases almost exponentially as DTE becomes smaller (closer to the event). Visually speaking, the distribution shape in Figure \ref{fig:DTE_all} is very similar to the function of $exp(1/x)$, therefore we convert the scale in $x$ axis from DTE to $log(DTE+1)$ \footnote{Since DTE can be smaller than 1, therefore we use $DTE+1$ to make everything positive.} as shown in Figure \ref{fig:DTE_all_log}. Now the values become more evenly distributed. We equally divide $log(DTE+1)$ into $L$ non-overlapping \textbf{bins} $BT = \{bt_1,bt_2,...,bt_L\}$  \footnote{We let the whole range of $BT$ cover $95\%$ of data, and all the data are added to their closest bin respectively.}.  Each time-bin $bt_j$ represents a certain range in $log(DTE+1)$ scale. 


From now on, we denote $G_k{(gs_i, bt_j)}$ as the ticket price of grid $gs_i$ at time bin $bt_j$ for event $k$, which is computed by the median of all the available transactions for the seats at grid $gs_i$ at time-bin $bt_j$. We denote $G_{k, bt_j} \in \mathbb{R}^{m \times 1} = \{G_k(gs_1, bt_j),\allowbreak G_k(gs_2, bt_j),\allowbreak ..., G_k(gs_m, bt_j) \} $, i.e. the median price collections of all the grids at $bt_j$. 

To conclude, this layer coarsens the seat-time to grid-bin resolution. It is similar to average pooling layer which is usually used in image modeling. However, there are two differences: 1) unlike pooling layer, it doesn't only compress spatial information but also temporal information;  2) the motivation of pooling layer is to reduce the model size, but our coarsening layer is to mitigate the spatio-temporal sparsity in the input data.


\subsection{Module 2: Spatial Modeling}
\label{sec:module2}

In Module 2, we employ a convolutional neural network (ConvNet) to explore the spatial dependency of ticket selling prices. 

First of all, we reorganize $G_{k, bt_j}$ into a two dimensional matrix with the first dimension as the row-wise coordinate and the second as the column-wise coordinate of grids (refer to Figure \ref{fig:auditorium-grid}). 

Then, we set the $G_{k, bt_j}$ as $\mathcal{X}_{k, bt_j}^{0}$, and  feed it to a number of convolutional layers. Assuming that $\mathcal{X}_{k, bt_j}^{\ell-1}$ denotes the feature maps in the $(\ell-1)$-th convolutional layer, the output of $\ell$-th layer is given by: 
\begin{align}
\label{eq:goal}
& \mathcal{X}_{k, bt_j}^{\ell} = \sigma_f(C^{\ell} * \mathcal{X}_{k, bt_j}^{\ell-1}+b^{\ell}),  
\end{align}  
where $*$ denotes the convolutional operation, $\sigma_f(\cdot)$ is the activation function of ReLU, $C^{\ell}$ denotes the convolutional filter and $b^{\ell}$ is a bias term at the $\ell$-th layer. The convolutional filters we used are three kernels of size $1\times 3$, three kernels of size $2\times 1$ and three kernels of size $2\times 3$, and they all have zero-padding. 
The output of the last convolutional layer is flattened to a vector, which we denote as $\bar{\mathcal{X}}_{k, bt_j} \in \mathbb{R}^{m \times 1}$. Then a fully connected layer is applied as below:  
\begin{align}
\label{eq:module2dense}
& D_{k, bt_j} = \sigma_f(W_{D} \bar{\mathcal{X}}_{k, bt_j}+b_{D}),  
\end{align}  
where $W_{D} \in \mathbb{R}^{m \times m}$ is a weight layer, and $b_D \in \mathbb{R}^{m \times 1}$ is a bias term.
The spatial dependency of grid prices is now captured by $D_{k, bt_j} \in \mathbb{R}^{m \times 1}$.


\subsection{Module 3: Temporal Modeling}
\label{sec:module3}
The spatial feature maps generated by Module $2$ is temporally dependent on the previous time bins. In Module $3$ we use recurrent neural network (RNN) to explore this temporal dependency. Specifically, we choose Gated Recurrent Units (GRUs) among all the RNN methods due to its comparable/better performance and less amount of parameters \cite{chung2014empirical,ugurlu2018electricity}.  

Given the output from Module $2$, we concatenate $D_{k, bt_j}$ with the event information $\mathcal{I}_{k}$. We denote the concatenated data as $\widetilde{D}_{k, bt_j} \in \mathbb{R}^{d \times 1}$, where $d = m + q$ and $q$ is dimensions of event information (such as event time, team's recent performance and so on) \footnote{We use OneHotEncoder to convert categorical features (if there is any) into numerical.}. 

Specifically, given the previous hidden state $\mathcal{H}_{k,bt_j-1} \in \mathbb{R}^{h \times 1}$, the current hidden state $\mathcal{H}_{k,bt_j}$ is updated with: 
\begin{align}
\label{eq:gru}
& \mathcal{H}_{k,bt_j} = GRU(\widetilde{D}_{k,bt_j}, \mathcal{H}_{k,bt_j-1}),  
\end{align}  
where the GRU cell \cite{chung2014empirical} is formulated as: 
\begin{align}
\begin{split} 
\label{eq:gru}
& z_{k,bt_j} = \sigma_g(W_{z}\widetilde{D}_{k,bt_j} + U_{z}\mathcal{H}_{k,bt_j-1}+b_{z}), \\
& r_{k,bt_j} = \sigma_g(W_{r}\widetilde{D}_{k,bt_j} + U_{r}\mathcal{H}_{k,bt_j-1}+b_{r}),\\
& \mathcal{H}_{k,bt_j} = (1-z_{k,bt_j})\circ \mathcal{H}_{k,bt_j-1} \\ &+ z_{k,bt_j} \circ \sigma_h(W_{h}\widetilde{D}_{k,bt_j} + U_{h}(r_{k,bt_j} \circ \mathcal{H}_{k,bt_j-1})+b_{z}), 
\end{split}
\end{align}
where $z_{k,bt_j} \in \mathbb{R}^{h \times 1}$ is the update gate vector, $r_{k,bt_j} \in \mathbb{R}^{h \times 1}$ is the reset gate vector, $\sigma_g(\cdot)$ is tanh function, $\sigma_h(\cdot)$ is sigmoid function, $W_{z}$, $W_{r}$  and $W_{h}$ are the weight matrices of size $h \times d$, $U_{z}$, $U_{r}$ and $U_{h}$ are the weight matrices of size $h \times h$, and $b_{z}$, $b_{r}$ and $b_{h}$ are the bias vectors of size $h \times 1$.     

We then feed the output of GRU at each time bin $\mathcal{H}_{k,bt_j}$ to a fully connected dense layer:
\begin{align}
\label{eq:module4dense}
& \widehat{G}_{k, bt_j} = W_{\widehat{G}} \mathcal{H}_{k,bt_j}+b_{\widehat{G}},  
\end{align}  
where $W_{\widehat{G}} \in \mathbb{R}^{m \times h}$ and $b_{\widehat{G}} \in \mathbb{R}^{m \times 1}$. So $\widehat{G}_{k, bt_j} \in \mathbb{R}^{m \times 1}$ is the output of Module $3$, which is the predicted grid price at time bin $bt_j$.

Essentially, Module $2$ and Module $3$ jointly model the spatial patterns of ticket transactions with temporal dependency.  

\subsection{Module 4: Data Refining}
\label{sec:module4}
In Module $4$, we expand the data from grid-level to seat-level. 
Firstly, we expand $\widehat{G}_{k, bt_j}\in \mathbb{R}^{m \times 1}$ to a vector of size $n \times 1$, where $n$ is the total number of seats. Each grid is expanded to the full size of its number of seats by duplicating the predicted grid price.  Then we concatenate the new vector with the corresponding seats' coordinates and date-to-event of the transactions column-wisely (as additional channels), and denote the whole matrix as $\widetilde{G}_{k, bt_j}  \in \mathbb{R}^{n \times e}$, where the columns include the predicted seat prices (expanded from $\widehat{G}_{k, bt_j}$), the date-to-event of transactions and the spatial coordinates of the seats. 

Finally, we apply a series of fully connected layers: 
\begin{align}
\label{eq:module4dense}
& \widehat{P}_{k, bt_j} = \sigma_f( \widetilde{G}_{k, bt_j}\widehat{W}_{\widehat{P}}+\widehat{b}_{\widehat{P}})\widetilde{W}_{\widehat{P}} +  \widetilde{b}_{\widehat{P}},  
\end{align}  
where $\widehat{W}_{\widehat{P}} \in \mathbb{R}^{e \times \gamma}$, $\widehat{b}_{\widehat{P}} \in \mathbb{R}^{n \times \gamma}$, $\widetilde{W}_{\widehat{P}} \in \mathbb{R}^{\gamma \times 1}$ and $\widetilde{b}_{\widehat{P}} \in \mathbb{R}^{n \times 1}$. 

\subsection{Our Loss Function}
\label{sec:loss}
Here we design a special bi-level loss function to integrate the two-level prediction output.  
The model has intermediate output $\widehat{G}_{k, bt_j}$ on grid-level prediction and final output  $\widehat{P}_{k, bt_j}$ on seat-level prediction. Although our final objective is to get better prediction of seat price, we found that combining the grid-level into loss function helps in getting better performance on the seat-level prediction. We will provide further analysis in the experiment section.  

Our final loss function consists of two parts, one for the coarsened targets (grid level), and one for the original targets (seat level): 

\begin{align}
\begin{split} 
\label{eq:loss}
& \mathcal{L} = \alpha \sum_{k,bt_j} ||{\mathbbm{1} _m(G_{k, bt_j} - \widehat{G}_{k, bt_j})}||_F^2 \\ 
&+ \beta \sum_{k,bt_j}||{\mathbbm{1} _n(P_{k,bt_j} - \widehat{P}_{k, bt_j})}||_F^2, 
\end{split}
\end{align} 
where 
\begin{align}
\label{eq:goal}
&\mathbbm{1}_{(i)} = 
	\begin{cases}
		0, & \text{if }\  \text{ corresponding ground-truth is missing}; \\
		1, & \text{otherwise}. 
	\end{cases}
\end{align}

The first part is to measure the residuals between the predicted and the actual grid price, while the second part is to measure the residuals between the predicted and the actual seat price. In the grid level, the sparsity is low and therefore most of the $\mathbbm{1} _m$ are $1$. But on the seat level, the ground truth is much more sparse and therefore only a small portion of data will be involved. Moreover, our final objective is to get better accuracy on the seat level prediction, therefore in practice we set $\beta > \alpha$. We employ Adam optimizer to minimize  $\mathcal{L}$.

%
%
%
%

%% file: Discussion.tex
\section{Related Work and Discussion}
\label{sec:discuss}
Price prediction is a great challenge due to the fact that it is an immensely complex, chaotic and dynamic problem. There are many studies from various areas aiming to take on that challenge and machine learning approaches have been the focus of many of them. Particularly, a consensus has been reached that price prediction is a highly nonlinear and time-variant problem \cite{adebiyi2014comparison}. 

Modern nonlinear machine learning methods have often been used in price prediction. 
For stock market price prediction \cite{patel2015predicting,khaidem2016predicting}, random forest is claimed to have better performance than Artificial Neural Network (ANN), Support Vector Machine (SVM), and naive-Bayes model. Specifically in the work of \cite{khaidem2016predicting}, random forest is proved to be robust in predicting future direction of stock movement.
On the other hand, Gradient Boosting Machine based methods like XGBoost have shown good performance on price prediction of crude oil, electricity and gold market \cite{pierdzioch2015forecasting,hong2016probabilistic,gumus2017crude}. It is popular because it has comparatively low variance and is able to recognize trends and fluctuations, and it can simultaneously handle variables of the nature of time indexes and static exogenous variables. 

Recently, Deep Neural Networks have been successfully used in many real world applications. Among all these methods, Recurrent Neural Networks (RNNs) have been
proposed to address time-dependent learning problems. In particular, Long Short Term Memory (LSTM) and Gated Recurrent Units (GRU) are tailor-made for time series price estimation \cite{nelson2017stock,chen2017stock,ugurlu2018electricity}. 

However, event ticket price prediction is very special compared with other price prediction problems, in the sense that the training data is extremely sparse. For each historical event, there is at most one transaction for any seat across time. And there are many timestamps that have no transaction available at all. The data sparsity is a deteriorating factor for all training based methods, but in particular for neural network based methods, of which performance heavily depends on the availability of large training data \cite{ugurlu2018electricity}. For example, Convolutional Neural Network (ConvNet/CNN) discovers the spatial dependency in the neighborhood content. But simply applying it to the sparse transaction data at each timestamp will lead to all zero weights due to the weight sharing problem. Therefore, although there are many recent research that explore deep learning in the direction of spatio-temporal prediction \cite{liu2018distribution,xu2018muscat,yao2018deep,yao2019revisiting,zhang2019flow}, none of them is applicable for event ticket price prediction due to the zero weighting problem.

%% file: Experiment.tex
\section{Experiment}
\label{sec:experiment}
In this section, we demonstrate the performance of our ETPP by a thorough comparison with a number of popular baselines and its variants on actual ticket transaction dataset.

\subsection{Experiment Setup}
\label{sec:experiment_setup}

\noindent\textbf{A Ticket Transaction Dataset from NBA} \\
Over the last couple of years, National Basketball Association (NBA) is growing in popularity. It is the most followed sports league on social media with more than 150 million followers. The average NBA ticket price for the $2018-2019$ season is up $14.01\%$ from the average ticket price of $\$78.00$ during the $2015-2016$ season. Therefore there is urgent need for any of the NBA teams to design a ticket price prediction model for successful promotional campaign. 

We use nine datasets to test our model. The data we use is from one NBA team's ticket transaction database. It involves all the available transactions from season $2014-2015$ to $2018-2019$. In total there are over $180,000$ transaction records from $246$ historical matches. Each record has the ticket sale price, seat location, and the information of the corresponding match. It is also worth to mention that the input and the price are all standardized during preprocessing. 

To simulate practical use, backtesting is used in the experiment to test our prediction accuracy. We build nine different datasets to backtest our model by selecting nine representative dates to separate the data. For each season we define three dates to represent starting of the season (Nov 1st), mid-season (Jan 1st), and late season (Feb 1st). In total we choose nine representative dates, from Nov 1st 2016 to Feb 1st 2019. For each representative date, we choose the $14$ matches right after that date as testing set, $10$ matches before that date as validation set, and all the data before the validation set for training.

\vspace{0.6em}
\noindent\textbf{Evaluation Metric} \\
We use Mean Squared Error (MSE) and Mean Absolute Percentage Error (MAPE) as the evaluation metrics in our experiment.
MSE is one of the most popular regression metrics. It is the second moment of the error, and thus incorporates both the variance of the estimator (how widely spread the estimates are from one data sample to another) and its bias (how far off the average estimated value is from the truth). We include it here since we apply MSE-like loss function in our model design, and for the other baselines we also use MSE as the loss function. 
MSE loss is defined as:
\begin{align}
\label{eq:mse}
& MSE = \frac{1}{T}\sum_{i}^{T}(p_i-\hat{p}_i)^2, 
\end{align} 
where $p$ ($\hat{p}$) is the actual (predicted) price of the transaction and $T$ is the total number of transactions. 

On the other hand, MAPE is included to account for the big differences in ticket price for different matches. This metric takes the relative errors of ticket price into consideration, since some ticket prices are in the thousands while some are as low as less than 100 dollars. Especially, the data doesn't contain any zero (there is no free ticket). So it is always meaningful to calculate MAPE. 
MAPE loss is defined as:
\begin{align}
\label{eq:mape}
& MAPE = \frac{1}{T}\sum_{i}^{T}|\frac{p_i-\hat{p}_i}{p_i}|, 
\end{align} 
where $p$ ($\hat{p}$) is the actual (predicted) price of the transaction and $T$ is the total number of transactions. 

Since not all the baselines we included generate grid-level output, we only compare the seat-level output which are available in all models. For each of the nine datasets, we run every method $10$ times and calculate the average MSE and MAPE. And we also measure the standard error of MSE and MAPE across the nine datasets for each method.

\vspace{0.6em}
\noindent\textbf{Baseline Methods} \\
We compare our ETPP method with the following six baselines. 

\begin{itemize}
\item \noindent{\em GameMedian:}  Tickets sales usually start long before the match date. Therefore we are able to use the median of the sold tickets of each game as a naive prediction model. 
\item \noindent{\em SectionMedian:} A slightly-improved but still naive model treats the median for loge and balcony seats separately. However, any further area segregation is difficult due to data sparsity.
\item \noindent{\em Linear Model (Linear):} It is a multiple linear regression approach for modeling the relationship between ticket prices, seat location and match information. But it treats transactions temporal-independently with each other.  
\item \noindent{\em Random Forest (RNFR) \cite{khaidem2016predicting}:} An ensemble technique with the use of multiple decision trees and Bootstrap Aggregation. It can successfully predict the price change direction in stock market \cite{khaidem2016predicting}. 
\item \noindent{\em XGBoost (XGB) \cite{chen2016xgboost}:}  An improved method based on popular Gradient Boosting Machine that has been applied on house price prediction \cite{Kaggle}.
\item \noindent{\em Gated Recurrent Unit Network (GRU) \cite{ugurlu2018electricity}:} GRU is a popular RNN model that has been used for price prediction \cite{ugurlu2018electricity}. For each game we concatenate all the ticket transactions to formulate a time series and feed these time series into the GRU model used in \cite{ugurlu2018electricity}. 
\end{itemize}

Besides comparing against popular baselines, we are also interested in verifying the necessity of the key components of our design. Therefore we include the following variants of ETPP in our comparison:  
\begin{itemize}
\item \noindent{\em $ETPP_1$:}  ETPP framework without Module $3$ (temporal modeling). 
\item \noindent{\em $ETPP_2$:}  ETPP framework without Module $2$ (spatial modeling).  
\item \noindent{\em $ETPP_3$:}  ETPP framework with seat-level loss function only (no grid-level loss). 
\item \noindent{\em $\mathbf{ETPP}$:}  The whole ETPP framework. 
\end{itemize}
The hyper-parameters of our ETPP are tuned according to the best performance on validation set. Specifically, the number of bins $L$ is set to $20$, the hidden dimensions ($h$) in GRU is set to $30$, $\gamma$ in Module $4$ is set to $7$, $\alpha$ and $\beta$ in our loss function (Equation \eqref{eq:loss}) are set to $0.3$ and $0.7$.  
For all the baselines, we tune the parameters according to validation set performance and to the best of our effort. 

\begin{figure}[!h]
\centering
\includegraphics[width=0.9\linewidth]{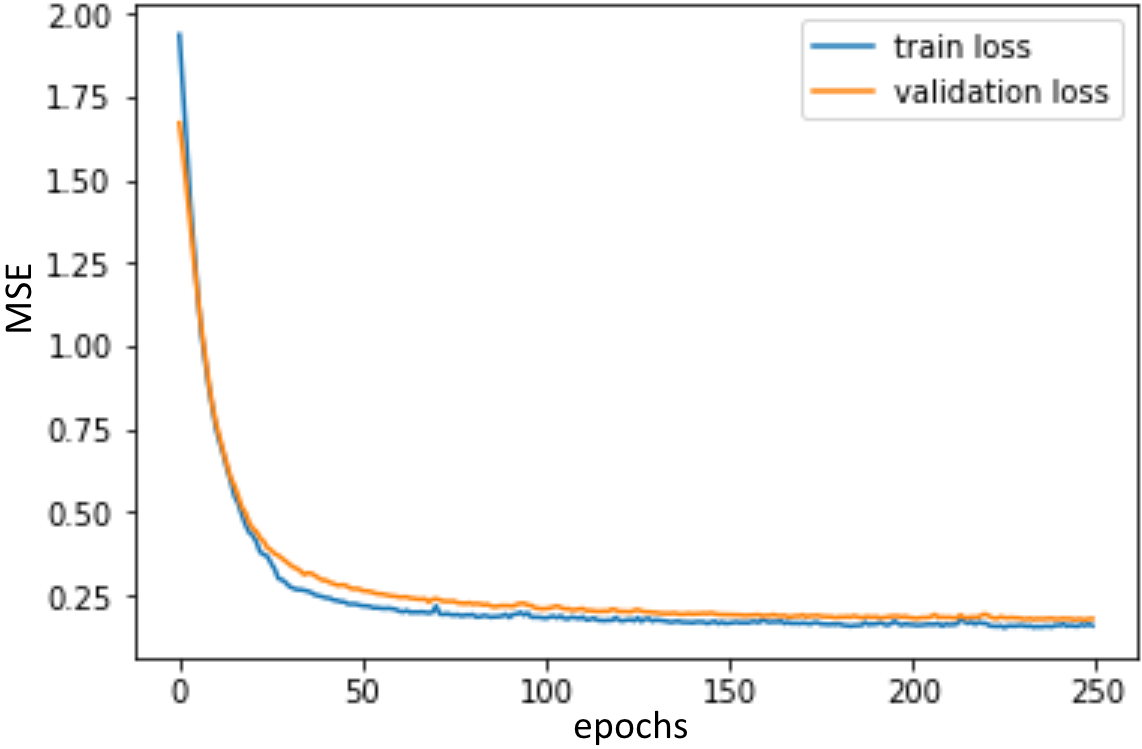}
 \caption{The convergence of our ETPP.}
 \label{fig:convergence}
\end{figure}

\subsection{Convergence of Our ETPP}
\label{sec:experiment_convergence}
Before comparing against different baselines, it is necessary to show the convergence of our ETPP method. We plot the training loss and validation loss across training epochs in Figure \ref{fig:convergence}. It shows that both the training loss and validation
loss converge well, proving that our ETPP model is reasonably stable.

\begin{figure*}[!h]
	\centering
	\subfigure[MSE and standard error comparison.]{\includegraphics[width=0.8\linewidth]{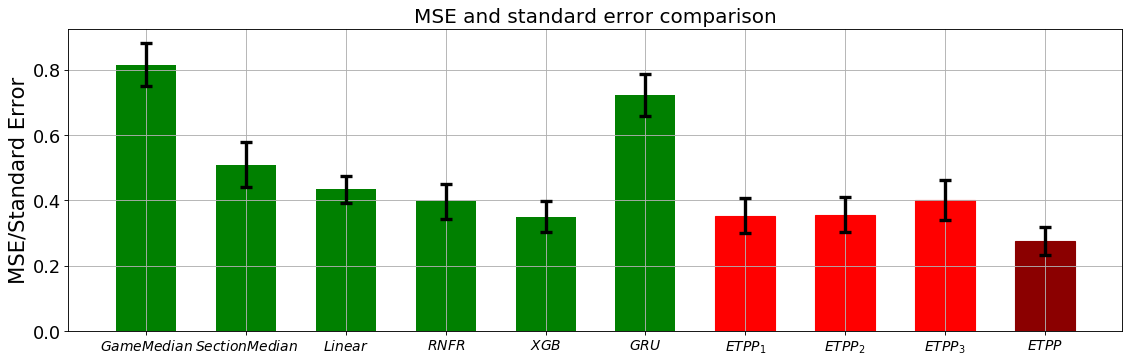} \label{fig:MSE}}
	\subfigure[MAPE and standard error comparison]{\includegraphics[width=0.8\linewidth]{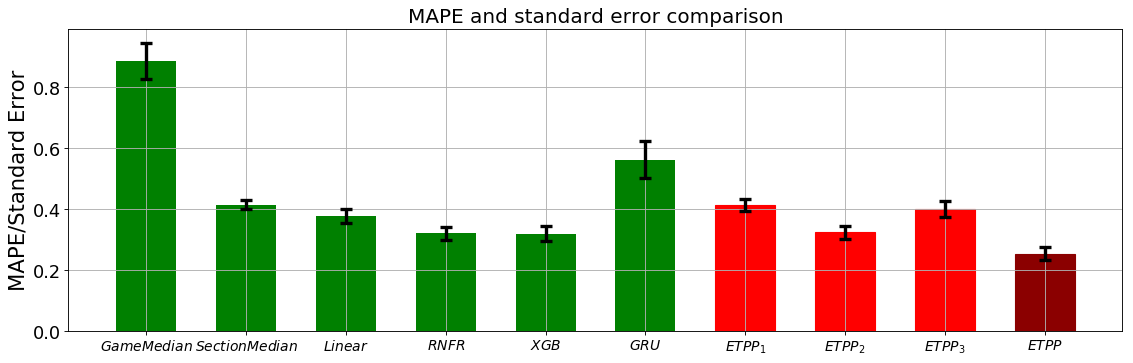} \label{fig:MAPE}}
	\caption{Result comparsion}
	\label{fig:full_result}
\end{figure*}

\subsection{Comparison of Prediction Performance}
\label{sec:experiment_comparison}
Figure \ref{fig:full_result} shows the full comparison result on the average MSE and MAPE together with the standard error on the nine datasets. We have the following observations: 
\begin{enumerate}[i)]
\item Obviously, our ETPP is superior to all the baselines, on both MSE and MAPE. Specifically, ETPP's MSE and MAPE is over $20\%$ lower than XGB, which performs the best among all baselines. Meanwhile, our standard error is also lower or comparable to the best performance from the other methods. 
\item It is not surprising that GameMedian has the worst performance, since it does not use any information from historical matches. SectionMedian improves a lot from the local spatial statistics in seat neighborhood. This proves that spatial dependency is one of the key factors in ticket price prediction. 
\item At first it may be surprising to see that popular RNN model like GRU fails miserably in this problem. But it can be explained by the curse of data sparsity. The concatenation of all the ticket transactions along time axis without any coarsening returns extremely sparse data. 
Besides, any two consecutive transactions are probably far away with each other in spatial position. RNN model is unable to explore the temporal dependency with such sparse and intermittent data. 
\item Among the traditional regression methods, nonlinear ones like Random Forest (RNFR) and XGBoosting (XGB) clearly outperform linear regression (Linear), which shows that the ticket prediction is (closer to) a nonlinear problem. XGB is slightly better than RNFR, which may be because that in terms of training objective, Boosted Trees(GBM) tries to add new trees that compliment the already built ones. This normally gives better accuracy with less trees. 
\item From the comparison between different variants of ETPP, we can clearly see that the two-level loss function, spatial modeling (ConvNet) and temporal modeling (GRU) all contribute to the superior performance of ETPP.
\end{enumerate}

\begin{figure}[!h]
\centering
\includegraphics[width=0.9\linewidth]{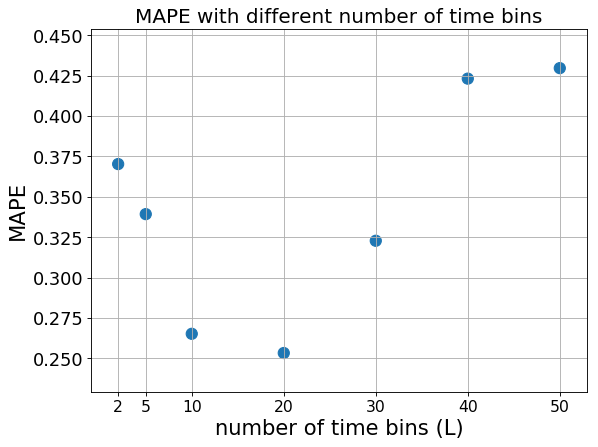}
 \caption{MAPE with different number of time bins.}
 \label{fig:MAPE_bin}
\end{figure}

\subsection{Effect of Different Coarsening Resolution}
\label{sec:experiment_bin}
In Section \ref{sec:module1} we describe the coarsening in both spatial and temporal dimensions. Here we discuss the effect of different coarsening resolution. 

While splitting space into grids is usually with prior knowledge or guidance (e.g. by the auditorium layout), there can be different options to split along the temporal dimensions. We already explained our strategy in Section \ref{sec:module1} of using log-scale in time axis. Here we want to show the effect of using different number of bins (L). Figure \ref{fig:MAPE_bin} shows MAPE of our ETPP with different values of $L$. The best result comes with the setting that is not too big or small. The reason is that: very small $L$ leads to much less sparsity but fewer samples in time series, which results in dropping useful information; On the other hand, setting $L$ too big increases sparsity and makes model unstable. In our experiment we find that $L=20$ gives the best prediction performance for the real-world dataset we used here.

%% file: Conclusion.tex
\section{Conclusion}
\label{sec:conclusion}
In this work, we target on solving the curse of sparsity in predicting event ticket sale price. We design a deep neural network framework which first coarsens the sparse input to dense format, then explores the spatial and temporal dependency, and finally expands the intermediate output back to the original format. Furthermore a bi-level loss function is proposed to get better prediction accuracy. Experiments on real world ticket transaction data proves that our method outperforms the popular baselines in price prediction.